\documentclass{article}

\PassOptionsToPackage{table}{xcolor}
\PassOptionsToPackage{numbers,sort&compress}{natbib}
\usepackage[preprint]{neurips_2026}

\usepackage[utf8]{inputenc}
\usepackage[T1]{fontenc}
\usepackage{hyperref}
\usepackage{url}
\usepackage{booktabs}
\usepackage{graphicx}
\usepackage{microtype}
\usepackage{xcolor}
\usepackage{wrapfig}

\usepackage{amsmath,amsfonts,bm}

\def\eqref#1{equation~\ref{#1}}

\def\1{\bm{1}}

\DeclareMathAlphabet{\mathsfit}{\encodingdefault}{\sfdefault}{m}{sl}
\SetMathAlphabet{\mathsfit}{bold}{\encodingdefault}{\sfdefault}{bx}{n}

\definecolor{darkblue}{rgb}{0, 0, 0.5}
\hypersetup{colorlinks=true, citecolor=darkblue, linkcolor=darkblue, urlcolor=darkblue}

\usepackage{pifont}
\newcommand{\cmark}{\ding{51}}              
\newcommand{\xmark}{\ding{55}}              
\newcommand{\pmark}{$\bigcirc$}             

\usepackage{enumitem}
\usepackage{xspace}
\newcommand{\ours}{ToolSciVer\xspace}

\title{\ours: Multimodal Scientific Claim Verification with Visual Tool Augmented Reinforcement Learning}

\author{
Binglin Zhou$^{1}$ \quad
Peng Shi$^{2}$ \quad
Ryo Kamoi$^{1}$ \quad
Nan Zhang$^{1}$ \quad
Rui Zhang$^{1}$\thanks{Corresponding author.} \\
$^{1}$The Pennsylvania State University \qquad
$^{2}$University of Waterloo \\
\texttt{\{bbz5169, rmz5227\}@psu.edu}
}

\begin{document}

\maketitle

\begin{abstract}

Multimodal Scientific Claim Verification (MSCV) requires models to verify scientific claims using visually grounded evidence from papers, including figures, tables, charts, and textual context. 
However, existing methods often fail because they struggle to locate decisive visual evidence, accurately read structured scientific visuals, and integrate multimodal observations into reliable reasoning. 
We introduce \ours{}, the first tool-augmented framework for MSCV to our knowledge. \ours{} equips a VLM with three type-aware visual tools, table row/column focus, chart-to-structure parsing, and high-resolution region zoom, which convert dense scientific visuals into explicit, claim-facing evidence, and trains the policy with Group Relative Policy Optimization (GRPO) under a composite reward of answer correctness, format validity, length control, tool-use efficiency, and tool-validity penalties. 
Experiments on \textsc{SciVer} and \textsc{MuSciClaims} datasets on five VLMs from three model families (Qwen, InternVL, Gemma) demonstrate that our method achieves superior performance compared to four competitive baselines including prompting-based and RL-based tool-use methods, highlighting the effectiveness of learned, type-aware tool use for scientific claim verification.\footnote{Code is available at \url{https://github.com/psunlpgroup/Tool-Sciver}.}

\end{abstract}

\section{Introduction}

Multimodal scientific claim verification (MSCV) is critical for building trustworthy scientific information systems~\citep{wang2025sciver,lal2025musciclaims}. In MSCV, a model determines whether a claim is supported or refuted by evidence that spans both textual descriptions and scientific visuals. Unlike traditional textual fact verification~\citep{thorne2018fever,wadden2020scifact,wadden2022multivers,wadden2022scifactopen,alvarez2024zeroshot}, MSCV requires integrating heterogeneous modalities with different structures by addressing two key challenges: extracting localized, structured visual evidence and performing multi-step reasoning over that evidence to verify the claim.

Despite growing interest, existing approaches to MSCV remain limited in addressing these challenges.
Recent studies highlight that frontier proprietary models, such as GPT, Claude, and Gemini, still struggle particularly when correct verification depends on precise visual interpretation with tables, charts, and figures~\citep{wang2025sciver,lal2025musciclaims,ho2026formatmatters,ho2026sciclaimeval}.
They often fail to localize correct visual evidence and struggle to aggregate information across modalities via reliable grounded multi-step reasoning~\citep{lal2025musciclaims,wang2025sciver,ho2026sciclaimeval}.
Meanwhile, while recent tool-augmented VLMs propose external visual operations for focused or extracted visual states \citep{fu2025refocus,wu2025vtoolr1,su2025openthinkimg}, these methods are developed for general-domain visual question answering and rely on a narrow repertoire of image-editing operations (e.g., cropping, zooming, highlighting), leaving them poorly suited to the claim-conditioned evidence localization and cross-figure aggregation over scientific tables, charts, and multi-panel figures that MSCV demands.
We summarize their limitations in Table~\ref{tab:method_comparison}.

\begin{figure*}[!t]
    \centering
    \includegraphics[width=\textwidth]{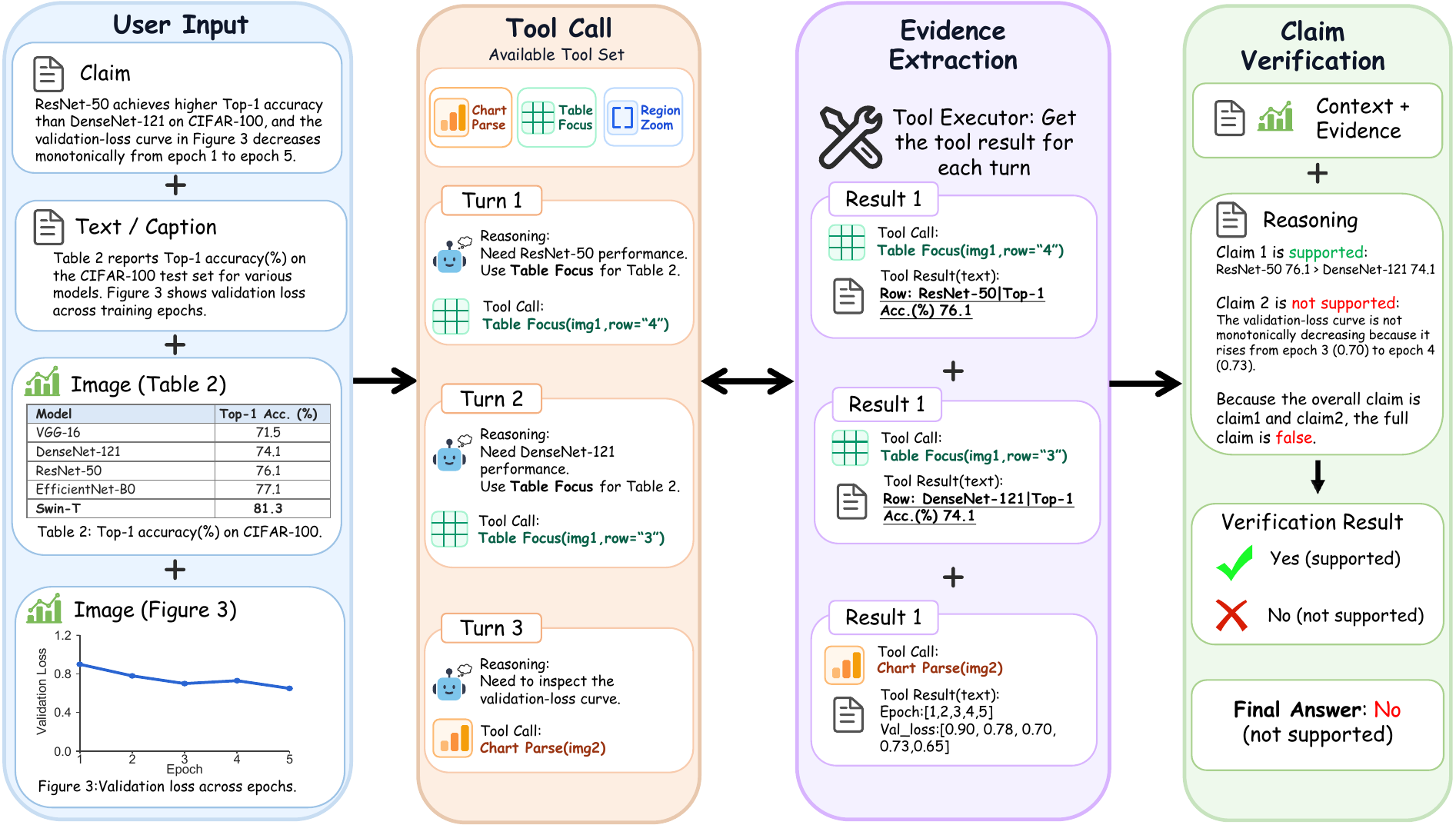}
    \caption{
    Overview of \ours framework for evidence-grounded tool-augmented scientific claim verification.
    Given a claim, textual context, and scientific visual evidence, the model can either answer directly or iteratively call type-aware visual tools to acquire claim-relevant evidence.
    The collected tool observations are added to the augmented context and used for final claim verification.
    }
    \vspace{-5mm}
    \label{fig:pipeline}
\end{figure*}

In this paper, we introduce \ours, a type-aware tool-augmented reinforcement learning framework for MSCV.
To address the above challenges, \ours includes two major innovations compared to prior work: (1) we design a suite of tools that are specialized to handle complex multimodal evidence in scientific papers, (2) we propose to use reinforcement learning to train models with a combination of rewards to efficiently use these tools and perform multi-step reasoning over extracted evidence. 
To be specific, as shown in Figure~\ref{fig:pipeline}, our framework equips a VLM with three specialized tools matched to the visual types most common in scientific evidence: an OCR-backed Table Focus tool that returns localized row or column evidence from tables, a Chart Parse tool that converts charts and plots into structured textual observations, and a Region Zoom tool that returns a high-resolution crop for general or multi-panel figures. 
Furthermore, effective use of these tools requires the model to jointly decide whether a tool is needed, which tool matches the visual type, what arguments identify the relevant evidence, and how to integrate the returned observation into its judgment.
Therefore, we train the model with Group Relative Policy Optimization (GRPO) \citep{shao2024deepseekmath} to learn selective, valid, and reliable tool use. Our reward combines final-answer correctness, output-format validity, length control, tool-use efficiency, and penalties for malformed or failed tool interactions, encouraging the model to acquire the right visual evidence only when needed.

To demonstrate the effectiveness of our method, we conduct a comprehensive set of experiments on two recent challenging benchmarks on MSCV, \textsc{SciVer}~\citep{wang2025sciver} and \textsc{MuSciClaims}~\citep{lal2025musciclaims}, which contains complex claims to verify with diverse evidence modality, scientific subdomains, and reasoning structures. 
We apply \ours on three backbone model familites (Qwen, InternVL, Gemma) and compare against four competitive baselines: non-tool CoT inference, prompt-only tool-use CoT inference, and two RL-based multimodal tool-use baselines \textsc{VTool-R1}~\citep{wu2025vtoolr1} and \textsc{OpenThinkIMG}~\citep{su2025openthinkimg}.
Across both benchmarks, \ours consistently improves over non-tool CoT inference and prompt-only tool access, and achieves stronger overall performance than the RL-based tool-use baselines.
Our further analyses and ablation study show that our method improves the performance by promoting more accurate visual evidence localization, more appropriate tool selection, and more reliable reasoning over extracted evidence across tables, charts, and general scientific figures.

Our contributions are summarized as follows:
\begin{itemize}[leftmargin=*,itemsep=0pt, topsep=0pt]
    \item We study Multimodal Scientific Claim Verification through the lens of visual evidence extraction and identify observation-stage visual evidence acquisition as a key bottleneck.
    \item We propose \ours, the first visual tool augmented framework for MSCV to the best of our knowledge. It consists of a type-aware visual tool suite that exposes multimodal, claim-relevant scientific evidence, trained with reinforcement learning for selective, valid, and efficient tool use.
    \item We empirically validate the effectiveness of \ours across challenging MSCV benchmarks in multiple scientific domains and reasoning difficulties, achieving state-of-the-art performance on different backbone model families compared with strong tool-use baselines.
\end{itemize}
\section{Related Work}

\begin{table}[t]
\centering
\small
\setlength{\tabcolsep}{4pt}
\renewcommand{\arraystretch}{1.1}
\resizebox{\textwidth}{!}{%
\begin{tabular}{l l ccc c c c}
\toprule
 & & \multicolumn{3}{c}{\textbf{Claim-Facing Visual Evidence Tools}} & & & \\
\cmidrule(lr){3-5}
\textbf{Method} & \textbf{Target Task} 
& \textbf{Table Evid.} 
& \textbf{Chart Evid.} 
& \textbf{Region Insp.} 
& \textbf{Struct. Obs.} 
& \textbf{Learning} 
& \textbf{Eff. Reward} \\
\midrule
VisProg~\citep{gupta2023visualprogramming}                  
& Gen. visual reasoning   
& \xmark & \xmark & \pmark & \xmark & Training-free & \xmark \\

ViperGPT~\citep{suris2023vipergpt}                
& Gen. visual reasoning   
& \xmark & \xmark & \pmark & \xmark & Training-free & \xmark \\

MM-ReAct~\citep{yang2023mmreact}                  
& Gen. multimodal reasoning 
& \xmark & \xmark & \pmark & \xmark & Prompt & \xmark \\

Visual Sketchpad~\citep{hu2024sketchpad}          
& Visual/math reasoning   
& \xmark & \xmark & \pmark & \xmark & Prompt & \xmark \\

ReFocus~\citep{fu2025refocus}                     
& Struct. img. underst.   
& \pmark & \pmark & \pmark & \xmark & Prompt/Sup. & \xmark \\

DePlot/MatCha~\citep{liu2023deplot,liu2023matcha} 
& Chart understanding     
& \xmark & \cmark & \xmark & \cmark & Sup./Pretrain & \xmark \\

VTool-R1~\citep{wu2025vtoolr1}                    
& Struct. visual reasoning 
& \pmark & \pmark & \pmark & \xmark & RL & \xmark \\

OpenThinkIMG~\citep{su2025openthinkimg}           
& Chart/tool reasoning    
& \xmark & \pmark & \pmark & \xmark & RL & \xmark \\

\midrule
\textbf{\ours{} (Ours)}                           
& \textbf{MSCV}           
& \cmark & \cmark & \cmark & \cmark & RL & \cmark \\
\bottomrule
\end{tabular}%
}
\caption{
Comparison of \ours{} with representative tool-augmented and visual reasoning methods.
\cmark{} denotes a dedicated claim-facing evidence extractor or parser for the corresponding visual type;
\pmark{} denotes partial or generic support through visual editing, sketching, cropping/zooming, OCR, or code-based visual tools, without a dedicated scientific-evidence extractor;
\xmark{} denotes no explicit support.
\textit{Struct. Obs.} refers to structured observations directly usable as verification evidence, such as table rows/columns or chart values, rather than raw OCR text, bounding boxes, or edited images.
}
\label{tab:method_comparison}
\end{table}

\paragraph{Scientific and multimodal claim verification.}
Claim verification is commonly formulated as evidence-grounded prediction, where a system retrieves evidence and predicts whether it supports or refutes a claim.
In the scientific domain, \textsc{SciFact} and follow-up work study verification against paper abstracts, full-document context, open-domain retrieval, and LLM-based zero-shot settings \citep{wadden2020scifact,wadden2022multivers,wadden2022scifactopen,alvarez2024zeroshot}.
Scientific table verification further shows that claims over scientific evidence require grounding, compositional reasoning, and careful treatment of ambiguity \citep{lu2023scitab}.
Recent benchmarks extend this line to multimodal evidence: \textsc{SciVer} evaluates multimodal scientific claim verification over scientific documents, while \textsc{MuSciClaims} focuses on figure-centric scientific claims \citep{wang2025sciver,lal2025musciclaims}.
Related work studies multi-hop multimodal claims, verification over visual and textual table representations, and unified multimodal fact verification \citep{wang2025mmcv,zhou2025m2tabfact,kishore2025multicheck}.
Recent cross-modal scientific verification studies also show that current multimodal models remain weak on table-, chart-, and figure-based evidence \citep{ho2026formatmatters,ho2026sciclaimeval}.
Our work builds on these benchmarks, but focuses on the mechanism behind the errors: acquiring claim-relevant visual evidence before making the final verification decision.

\paragraph{Visual evidence extraction from structured scientific visuals.}
Structured visual understanding has been studied for tables, charts, plots, and document-like images.
For tabular evidence, TabFact, SciTab, and M$^{2}$-TabFact show that table-based verification requires accurate grounding and compositional reasoning \citep{chen2020tabfact,lu2023scitab,zhou2025m2tabfact}.
For charts and plots, ChartQA highlights visual and logical reasoning over chart content, while DePlot and MatCha show the value of converting charts into explicit intermediate representations through plot-to-table translation or chart derendering \citep{masry2022chartqa,liu2023deplot,liu2023matcha}.
These works motivate our use of structured evidence, but fixed OCR, parsing, or cropping alone does not decide whether a tool is needed, which tool to call, or which row, column, chart element, or region is relevant to the claim.
We therefore treat visual evidence extraction as part of a learned verification policy rather than as fixed preprocessing.

\paragraph{Tool-augmented VLMs and reinforcement learning.}
Tool use has been widely studied for language models.
ReAct interleaves reasoning and acting through prompting, while Toolformer learns API use from self-supervised signals \citep{yao2023react,schick2023toolformer}.
Visual programming and tool-use systems such as VISPROG, ViperGPT, and MM-ReAct call external vision modules or generate executable programs without task-specific training \citep{gupta2023visualprogramming,suris2023vipergpt,yang2023mmreact}.
Visual Sketchpad and ReFocus further show that intermediate visual operations can expose focused visual states for multimodal reasoning \citep{hu2024sketchpad,fu2025refocus}.
Closest to our work, \textsc{VTool-R1} and \textsc{OpenThinkIMG} train VLMs to use visual tools with reinforcement learning \citep{wu2025vtoolr1,su2025openthinkimg}, while broader tool-learning work emphasizes reward design, execution reliability, and action efficiency \citep{qian2025toolrl,jiang2025verltool,wang2025actingless}.
Unlike these general visual tool-use settings, our method targets MSCV specifically: tools are organized by scientific visual type, their outputs are claim-facing evidence, and GRPO training encourages selective, valid, and efficient tool use for scientific verification.
\section{Our Method}
\label{sec:method}

\subsection{Task Formulation}

We study multimodal scientific claim verification (MSCV), where a model verifies a scientific claim using textual context and scientific visual evidence.
Each instance is denoted as $x=(q,\mathcal{V},t)$, where $q$ is the claim, $\mathcal{V}=\{v_i\}_{i=1}^{N}$ is the set of associated scientific visuals, and $t$ denotes optional textual context, such as captions, surrounding paragraphs, or benchmark metadata.
The model outputs a normalized prediction $\hat{y}$ in the benchmark label space.
In our experiments, both benchmarks are cast into binary verification with $\mathcal{Y}=\{\text{supported}, \text{refuted}\}$.

\subsection{Framework Overview}

The key challenge in MSCV is that decisive evidence is often visually localized and structurally constrained, requiring the model to read specific table entries, recover chart trends, or inspect local regions of dense scientific figures.
We therefore formulate MSCV as an evidence acquisition problem and propose \ours, a tool-augmented framework that selectively extracts claim-relevant visual evidence before making the final verification decision.

As shown in Figure~\ref{fig:pipeline}, given a claim, textual context, and scientific visual evidence, the policy either answers directly from the original input or issues a type-aware visual tool call.
A scheduler validates and executes each call, appends the returned observation to the context, and allows the model to continue reasoning over the augmented evidence.
The resulting trajectory is written as
$$
    \tau=(u_1,o_1,u_2,o_2,\ldots,u_T,\hat{y}),
$$
where $u_t$ denotes a model turn, $o_t$ denotes an optional tool observation, and $\hat{y}$ is the final answer.
Each turn contains either one tool call or the final answer, which keeps the action space narrow and grounds the interaction in visual evidence extraction rather than unrestricted agentic search.

This design follows three principles: tools should expose claim-relevant evidence rather than generic image descriptions; tool use should be type-aware across tables, charts, and general figures; and tool calls should be selective, invoked only when the original input is insufficient for reliable verification.

\subsection{Type-Aware Visual Tool Suite}

We design a compact tool suite for three common evidence structures in MSCV: tables, charts, and general scientific figures.
Each tool call targets one image through an \texttt{image\_path} argument, and the policy decides which tool to call and what arguments to provide.
Detailed tool interfaces are provided in Appendix~\ref{app:tool_interfaces}.\\
\textbf{Table Focus.}
Scientific claims often depend on localized tabular evidence, such as a method's performance on a specific dataset or the value of a particular metric.
For table images, we first convert the image into a structured CSV-like representation using OCR.
We then expose \texttt{focus\_row} and \texttt{focus\_column}, which return only the selected row or column.
This reduces irrelevant context and makes the returned evidence easier to compare against the claim.\\
\textbf{Chart Parse.}
Claims over charts or plots often require recovering trends, relative ordering, thresholds, or numerical values.
For chart images, \texttt{parse\_content} converts visual chart content into structured textual observations by using stronger VLM, such as JSON-like entries of chart elements and values.\\
\textbf{Region Zoom.}
Many scientific figures are dense, multi-panel, or diagrammatic, and the decisive evidence may occupy only a small local region.
For these cases, \texttt{image\_zoom\_in} crops and enlarges a model-selected region, returning a zoomed-in image.
Coordinates are normalized to a $[0,1000]$ scale, making the interface resolution-invariant.

\subsection{Reward Design}

Our reward aligns the policy with the deployment goal of MSCV: produce a correct final decision, follow the required output format, avoid overlong responses, and use tools only when they help acquire relevant visual evidence.
For a rollout $\tau$, we define
\begin{equation}
\label{eq:reward}
R(\tau)
=
s_{\text{eff}}(\tau)
\left(
r_{\text{ans}}(\tau)
+
\lambda_{\text{fmt}} r_{\text{fmt}}(\tau)
\right)
+
\lambda_{\text{len}} r_{\text{len}}(\tau)
+
\lambda_{\text{tool}} r_{\text{tool}}(\tau)
\end{equation}
In our experiments, we set $\lambda_{\text{fmt}}=\lambda_{\text{len}}=\lambda_{\text{tool}}=0.1$.

\textbf{Answer reward.}
The answer reward measures final task correctness:
$r_{\text{ans}}(\tau)=\mathbb{I}[\hat{y}=y]$
where $y$ is the ground-truth label and $\hat{y}$ is the normalized model prediction.\\
\textbf{Format reward.}
The format reward encourages the model to produce a parseable final response: 
$
r_{\text{fmt}}(\tau)
=
0.5\,I_{\text{reason}}(\tau)
+
0.5\,I_{\text{answer}}(\tau),
$
where $I_{\text{reason}}(\tau)$ indicates whether the final response contains the required reasoning segment under the prompt template, and $I_{\text{answer}}(\tau)$ indicates whether the response contains exactly one valid normalized final-answer marker.
This term stabilizes training and ensures that final predictions can be reliably extracted.\\
\textbf{Tool-efficiency coefficient.}
To encourage selective rather than excessive tool use, we introduce an OTC-style tool-efficiency coefficient.
Let $m(\tau)$ be the number of tool calls in rollout $\tau$.
Let $\mathcal{C}(x)$ be the set of strict-correct rollouts in the GRPO group, where a rollout is strict-correct if it predicts the correct label and ends with a valid final answer.
If $\mathcal{C}(x)$ is non-empty, we estimate the group-local optimal tool budget as $n^\star(x)=\min_{\tau_i\in\mathcal{C}(x)} m(\tau_i)$. 
If no strict-correct rollout exists, we set $s_{\text{eff}}(\tau)=1$.
Otherwise, for maximum tool budget $c_{\max}$,
\begin{equation}
s_{\text{eff}}(\tau)=
\begin{cases}
1, & n^\star=0,\; m=0, \\
\cos\left(\frac{m\pi}{2m+c_{\max}}\right),
& n^\star=0,\; m>0, \\
0, & n^\star>0,\; m=0, \\
\sin\left(\frac{m\pi}{m+n^\star}\right),
& n^\star>0,\; m>0.
\end{cases}
\end{equation}
This coefficient rewards correct trajectories that use close to the group-local minimal number of tool calls, while penalizing unnecessary tool use when a direct answer is sufficient.
We set $c_{\max}=5$.\\
\textbf{Length reward.}
We use a soft overlong penalty to discourage unnecessarily long generations.
Let $L(\tau)$ denote the completion length in tokens, let $L_{\max}$ be the maximum completion length, and let $C$ denote the soft-cache length.
We define the soft threshold as
$
L_{\text{soft}}=L_{\max}-C.
$
The length reward is
$
r_{\text{len}}(\tau)
=
-\frac{\max(L(\tau)-L_{\text{soft}},0)}{C}.
$
Thus, the penalty is zero below the soft threshold and decreases linearly as the response approaches the hard length limit.
In our experiments, $L_{\max}=4096$ and $C=768$.\\
\textbf{Tool-interaction penalty.}
The tool-interaction penalty discourages malformed tool turns and execution failures.
Let $e_{\text{turn}}(\tau)$ be the number of malformed tool turns, let $e_{\text{exec}}(\tau)$ be the number of tool-execution failures, and let $I_{\text{bad-end}}(\tau)$ indicate whether the rollout contains at least one tool-related error and still fails to terminate with a valid final answer.
We compute
\begin{equation}
\begin{aligned}
\tilde{r}_{\text{tool}}(\tau)=&
-0.20\,\mathbb{I}[e_{\text{turn}}\ge 1]
-0.10\,\max(e_{\text{turn}}-1,0) \\
&-0.30\,\mathbb{I}[e_{\text{exec}}\ge 1]
-0.15\,\max(e_{\text{exec}}-1,0) \\
&-0.40\,I_{\text{bad-end}}(\tau),
\end{aligned}
\end{equation}
and clip the result:
$
r_{\text{tool}}(\tau)
=
\operatorname{clip}_{[-1.5,0]}
\left(\tilde{r}_{\text{tool}}(\tau)\right).
$
This gives the model dense feedback for invalid tool behavior, rather than treating all tool failures as a single sparse error.


\subsection{Learning Selective Tool Use with GRPO}

Prompt-only tool use is brittle because the model must decide whether a tool is needed, select the appropriate tool family, provide valid arguments, and use the returned observation in the final decision.
To learn this behavior, we train the policy with Group Relative Policy Optimization (GRPO) \citep{shao2024deepseekmath}.
For each input $x$, we sample a group of $G$ rollouts $\mathcal{G}(x)=\{\tau_i\}_{i=1}^{G}$ from the current policy. 
After rollout completion, we assign a scalar reward $R_i$ based on equation~\ref{eq:reward} and compute the group-relative advantage
$
A_i=
\frac{R_i-\operatorname{mean}_{j=1}^{G}R_j}
{\operatorname{std}_{j=1}^{G}R_j}.
$
The policy is then optimized with the standard GRPO objective, encouraging high-reward trajectories while staying close to the reference policy.
\begin{equation}
\mathcal{J}_{\text{GRPO}}(\theta)
=
\frac{1}{G}
\sum_{i=1}^{G}
\sum_{t}
\min
\left(
r_{i,t}(\theta) A_i,
\operatorname{clip}(r_{i,t}(\theta),1-\epsilon,1+\epsilon)A_i
\right)
-
\beta D_{\mathrm{KL}}(\pi_{\theta}\|\pi_{\mathrm{ref}})
\end{equation}
$r_{i,t}(\theta)$ is the token-level policy ratio, $\pi_{\mathrm{ref}}$ is the reference policy, and $\beta$ controls the KL penalty.
\section{Experimental Setup}
\label{sec:experimental_setup}

\paragraph{Datasets.}
We evaluate on two MSCV benchmarks: \textsc{SciVer} \citep{wang2025sciver} and \textsc{MuSciClaims} \citep{lal2025musciclaims}.
\textsc{SciVer} contains scientific claims paired with multimodal evidence from scientific documents and reports four reasoning subsets: \emph{Analytic}, \emph{Direct}, \emph{Parallel}, and \emph{Sequential}.
\textsc{MuSciClaims} contains figure-centric scientific claims from biology, physics, and chemistry.
We use the binary verification setting for both datasets and report subset/domain accuracy together with sample-weighted overall accuracy.
Exact train/test split sizes are reported in Appendix~\ref{app:data_distribution}.

\paragraph{Models.}
We evaluate five open-source VLM backbones from three model families:
Qwen3.5-4B and Qwen3.5-9B \citep{qwen2026qwen35},
InternVL3.5-4B and InternVL3.5-8B \citep{wang2025internvl35},
and Gemma4-E4B \citep{farabet2026gemma4}.
All trainable methods are initialized from the same backbone and trained on the same MSCV training splits.

\paragraph{Baselines.}
We compare our method with four competitive baselines.
\textbf{Non-tool CoT inference}~\citep{wei2022chain} prompts the model to reason step by step over the claim, context, caption, and image, without access to external tools.
\textbf{Prompt-only tool-use CoT inference}~\citep{trivedi2023interleaving} gives the model access to the same MSCV tool APIs and scheduler as our method, but without tool-use training.
\textbf{Dataset-adapted \textsc{VTool-R1}}~\citep{wu2025vtoolr1}. and \textbf{dataset-adapted \textsc{OpenThinkIMG}}~\citep{su2025openthinkimg} retain the original tool suites and reward designs of \textsc{VTool-R1} and \textsc{OpenThinkIMG}, respectively, but we train them on the same MSCV training splits as our method. 
These baselines test whether general trained visual tool-use recipes are sufficient for MSCV.
Finally, we also report advanced proprietary models, including \textbf{GPT-4o}, \textbf{GPT-5.4}, and \textbf{Claude Sonnet 4.6}, as external references under the non-tool CoT setting; they are not part of the controlled tool-training comparison.

\section{Results and Analysis}
\label{sec:results}

We organize our results and analysis around the central question of whether \ours improves MSCV by helping VLMs acquire and use claim-relevant visual evidence.
Section~\ref{sec:main_results} first reports the main MSCV results across datasets, model families, and baselines.
To better understand why our method works, we conducted four diagnostic analyses on Qwen3.5-4B.
\begin{itemize}[leftmargin=*,itemsep=0pt, topsep=0pt]
    \item \textbf{Evidence acquisition} (Section~\ref{sec:rear_analysis}): Does the model obtain more claim-relevant visual evidence?
    \item \textbf{Tool selection} (Section~\ref{sec:type_aware_tool_selection}): Does the model learn to select tools according to the image type?
    \item \textbf{Efficiency} (Section~\ref{sec:efficient_tool_use}): Does the model solve MSCV with fewer tool calls and shorter responses?
    \item \textbf{Evidence-based reasoning} (Section~\ref{sec:reasoning_over_evidence}): Once evidence is acquired, does the model reason more reliably over it?
\end{itemize}

\subsection{Main Results}
\label{sec:main_results}


\begin{table*}[t]
\centering
\footnotesize
\renewcommand{\arraystretch}{1.05}
\setlength{\tabcolsep}{3.8pt}
\resizebox{\textwidth}{!}{
\begin{tabular}{llccccccccc}
\toprule
\multicolumn{2}{c}{\textbf{Model}} & \multicolumn{5}{c}{\textbf{\textsc{SciVer}}} & \multicolumn{4}{c}{\textbf{\textsc{MuSciClaims}}} \\
\cmidrule(lr){1-2} \cmidrule(lr){3-7} \cmidrule(lr){8-11}
\textbf{Backbone} & \textbf{Method} & \textbf{Analytic} & \textbf{Direct} & \textbf{Parallel} & \textbf{Sequential} & \textbf{Overall} & \textbf{Bio} & \textbf{Phy} & \textbf{Chem} & \textbf{Overall} \\
\midrule
\rowcolor{gray!18}
\multicolumn{11}{c}{\textbf{\textsc{Open-source Models}}} \\
Qwen3.5-4B & \textsc{CoT} & 71.88 & 74.19 & 68.57 & 65.74 & 70.10 & 60.46 & 51.04 & 71.84 & 60.99 \\
Qwen3.5-4B & Tool Prompt & 70.31 & 75.40 & 65.31 & 69.32 & 70.10 & 74.84 & 64.58 & 69.90 & 71.88 \\
Qwen3.5-4B & \textsc{VTool-R1} & 83.98 & \textbf{84.27} & \textbf{77.14} & 79.28 & \textbf{81.20} & 77.12 & \textbf{71.88} & \textbf{80.58} & 76.83 \\
Qwen3.5-4B & \textsc{OpenThinkIMG} & 73.44 & 77.42 & 74.29 & 78.09 & 75.80 & 77.78 & 68.75 & 77.67 & 76.04 \\
Qwen3.5-4B & Ours & \textbf{85.16} & 80.65 & \textbf{77.14} & \textbf{81.67} & \textbf{81.20} & \textbf{81.70} & 67.71 & 79.61 & \textbf{78.61} \\
\midrule
Qwen3.5-9B & \textsc{CoT} & 79.30 & 81.05 & 66.53 & 72.11 & 74.80 & 72.88 & 60.42 & 72.82 & 70.50 \\
Qwen3.5-9B & Tool Prompt & 70.70 & 70.97 & 62.04 & 63.35 & 66.80 & 79.41 & 67.71 & 79.61 & 77.23 \\
Qwen3.5-9B & \textsc{VTool-R1} & 77.34 & 84.27 & 70.61 & 76.89 & 77.30 & \textbf{80.07} & 68.75 & 83.50 & 78.61 \\
Qwen3.5-9B & \textsc{OpenThinkIMG} & \textbf{86.33} & 85.08 & \textbf{80.82} & \textbf{85.26} & \textbf{84.40} & 76.47 & \textbf{70.83} & \textbf{87.38} & 77.62 \\
Qwen3.5-9B & Ours & 76.56 & \textbf{87.90} & 77.14 & 78.49 & 80.00 & \textbf{80.07} & \textbf{70.83} & 82.52 & \textbf{78.81} \\
\midrule
InternVL3.5-4B & \textsc{CoT} & 69.64 & 65.74 & 63.76 & 66.33 & 66.37 & 69.93 & 53.13 & 63.11 & 65.35 \\
InternVL3.5-4B & Tool Prompt & 71.88 & 63.31 & 60.41 & 66.53 & 65.60 & 68.63 & 67.71 & \textbf{75.73} & 69.90 \\
InternVL3.5-4B & \textsc{VTool-R1} & \textbf{79.30} & 69.76 & 66.53 & 70.92 & 71.70 & 77.78 & 67.71 & \textbf{75.73} & 75.45 \\
InternVL3.5-4B & \textsc{OpenThinkIMG} & 71.48 & 67.74 & 66.12 & 68.53 & 68.50 & 74.51 & 68.75 & 73.79 & 73.27 \\
InternVL3.5-4B & Ours & \textbf{79.30} & \textbf{72.18} & \textbf{68.16} & \textbf{71.71} & \textbf{72.90} & \textbf{78.76} & \textbf{72.92} & 73.79 & \textbf{76.63} \\
\midrule
InternVL3.5-8B & \textsc{CoT} & 68.02 & 71.71 & 63.37 & 66.73 & 67.44 & 68.63 & \textbf{68.75} & 69.90 & 68.91 \\
InternVL3.5-8B & Tool Prompt & 68.75 & 60.08 & 55.51 & 55.38 & 60.00 & 69.61 & 61.46 & 67.96 & 67.72 \\
InternVL3.5-8B & \textsc{VTool-R1} & 73.83 & 67.34 & 62.04 & 63.35 & 66.70 & 75.82 & 67.71 & 63.11 & 71.68 \\
InternVL3.5-8B & \textsc{OpenThinkIMG} & 72.27 & 62.10 & 60.41 & 60.16 & 63.80 & 73.86 & 67.71 & \textbf{73.79} & 72.67 \\
InternVL3.5-8B & Ours & \textbf{79.69} & \textbf{74.19} & \textbf{73.47} & \textbf{78.09} & \textbf{76.40} & \textbf{78.43} & 66.67 & 71.84 & \textbf{74.85} \\
\midrule
Gemma4-E4B & \textsc{CoT} & 78.13 & 73.79 & \textbf{72.24} & 71.31 & 73.90 & 76.14 & 68.75 & 70.87 & 73.66 \\
Gemma4-E4B & Tool Prompt & 65.46 & 71.98 & 64.27 & 61.75 & 65.85 & 68.75 & \textbf{73.24} & 65.43 & 68.93 \\
Gemma4-E4B & \textsc{VTool-R1} & 78.91 & \textbf{77.82} & 69.39 & 72.91 & 74.80 & 68.95 & 56.25 & \textbf{77.67} & 68.32 \\
Gemma4-E4B & \textsc{OpenThinkIMG} & 81.25 & 76.21 & 62.04 & 62.95 & 70.70 & 72.22 & 70.83 & 73.79 & 72.28 \\
Gemma4-E4B & Ours & \textbf{82.30} & 72.65 & 66.22 & \textbf{84.29} & \textbf{76.47} & \textbf{76.19} & 69.88 & 73.26 & \textbf{74.25} \\
\midrule
\rowcolor{gray!18}
\multicolumn{11}{c}{\textbf{\textsc{Proprietary Models}}} \\
GPT-4o & \textsc{CoT} & 71.20 & 77.00 & 73.60 & 73.80 & 73.89 & 75.49 & 67.71 & 71.84 & 73.27 \\
GPT-5.4 & \textsc{CoT} & 53.52 & 75.00 & 66.53 & 56.97 & 62.90 & 78.76 & 68.75 & 70.87 & 75.25 \\
Claude Sonnet 4.6 & \textsc{CoT} & 71.48 & 79.44 & 65.31 & 65.74 & 70.50 & 83.01 & 72.92 & 74.76 & 79.41 \\
\bottomrule
\end{tabular}
}
\caption{
Main results on \textsc{SciVer} and \textsc{MuSciClaims}.
Accuracy (\%) is reported on \textsc{SciVer} reasoning subsets and \textsc{MuSciClaims} domain subsets, with sample-weighted Overall scores.
Open-source rows form controlled comparisons, while proprietary rows serve as external \textsc{CoT} references.
Bold marks the best score within each backbone block.
}
\label{tab:main_results}
\end{table*}

Table~\ref{tab:main_results} shows three main trends.
First, \ours improves over non-tool \textsc{CoT} on both benchmarks across all five open-source backbones, with particularly clear gains on the information-dense \textsc{MuSciClaims} benchmark.
Second, prompt-only tool access is unstable: on \textsc{SciVer}, Tool Prompt underperforms non-tool \textsc{CoT} for four of five backbones, showing that simply exposing tools does not reliably improve verification.
Third, \ours matches or improves over \textsc{VTool-R1} in all completed overall comparisons, including the Qwen3.5-4B tie on \textsc{SciVer}, and outperforms \textsc{OpenThinkIMG} on \textsc{MuSciClaims} and all completed \textsc{SciVer} settings except Qwen3.5-9B.
These results suggest that MSCV benefits from task-aligned visual-evidence tools beyond general visual tool-use training.

The proprietary \textsc{CoT} references further show that advanced closed-source VLMs are not uniformly reliable on MSCV: Claude Sonnet 4.6 performs best among them on \textsc{MuSciClaims}, whereas GPT-5.4 is substantially weaker on \textsc{SciVer}.
This underscores the difficulty of MSCV and motivates explicit visual-evidence acquisition for open-source VLMs.

\subsection{Claim-Relevant Visual Evidence Acquisition}
\label{sec:rear_analysis}

\begin{wraptable}[9]{r}{0.50\linewidth}
\vspace{-8pt}
\centering
\footnotesize
\setlength{\tabcolsep}{2.6pt}
\renewcommand{\arraystretch}{1.04}
\begin{tabular}{lcccc}
\toprule
\textbf{Method} & \textbf{Table} & \textbf{Chart} & \textbf{M-panel} & \textbf{Avg.} \\
\midrule
\textsc{CoT} & 93.00 & 87.00 & 69.00 & 83.00 \\
Tool Prompt & 87.00 & 88.00 & 75.00 & 83.33 \\
Ours & \textbf{98.00} & \textbf{93.00} & \textbf{79.00} & \textbf{90.00} \\
\bottomrule
\end{tabular}
\caption{Relevant Evidence Acquisition Rate (REAR, \%).}
\label{tab:rear}
\vspace{-10pt}
\end{wraptable}

\begin{figure*}[!b]
\centering
\begingroup
\definecolor{casepanelbg}{HTML}{F8FAFC}
\definecolor{caseborder}{HTML}{CBD5E1}
\definecolor{casered}{HTML}{B91C1C}
\definecolor{caseredbg}{HTML}{FEF2F2}
\definecolor{casegreen}{HTML}{047857}
\definecolor{casegreenbg}{HTML}{ECFDF5}
\definecolor{casecodebg}{HTML}{F1F5F9}

\newcommand{\caseinnerbox}[3]{%
  \begingroup
  \setlength{\fboxsep}{4pt}%
  \fcolorbox{#1}{#2}{%
    \begin{minipage}{\dimexpr\linewidth-2\fboxsep-2\fboxrule\relax}
    #3
    \end{minipage}}%
  \endgroup}

\newcommand{\casepanel}[2]{%
  \begingroup
  \setlength{\fboxsep}{6pt}%
  \fcolorbox{caseborder}{casepanelbg}{%
    \begin{minipage}[t][4.85cm][t]{#1}
    \raggedright
    #2
    \end{minipage}}%
  \endgroup}

\scriptsize
\casepanel{0.285\textwidth}{
  {\bfseries Input: chart + claim}\par
  \vspace{2pt}
  Claim asks for country-level evidence: \textbf{China (13)} and \textbf{Russia (10)} sanctioned BTC entities.\par
  \vspace{5pt}
  \includegraphics[width=\linewidth]{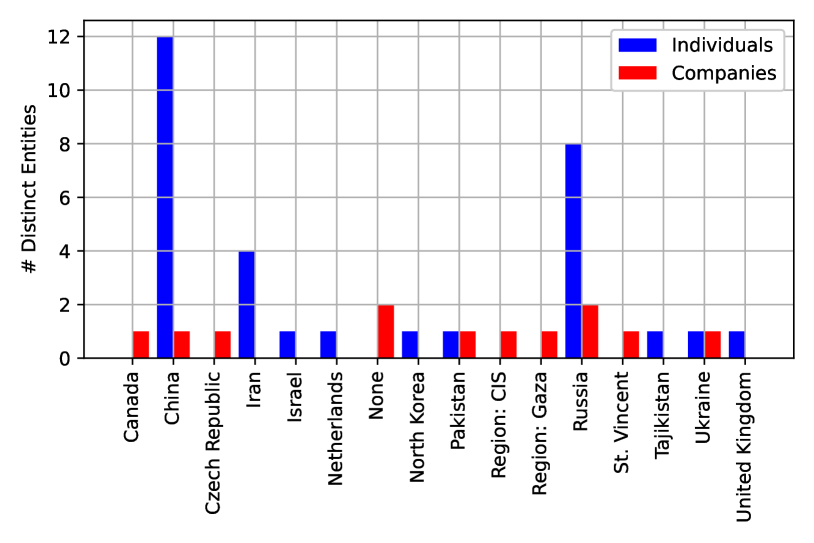}
}
\hfill
\casepanel{0.285\textwidth}{
  {\bfseries\textcolor{casered}{CoT Baseline}}\par
  \vspace{2pt}
  Notices large bars, but fails to bind them to labels and exact values.\par
  \vspace{5pt}
  \caseinnerbox{caseborder}{white}{
    \textcolor{gray}{``x-axis labels are cut off or not visible...''}\par
    \vspace{3pt}
    \textcolor{gray}{``China'' or ``Russia'' are not explicitly labeled.}\par
    \vspace{3pt}
    \textcolor{gray}{Evidence is the tall blue bars.}
  }\par
  \vspace{5pt}
  \caseinnerbox{casered}{caseredbg}{
    {\bfseries\textcolor{casered}{Missing evidence:}}\par
    \texttt{country labels + series values}\par
    \texttt{China: 12 + 1}\par
    \texttt{Russia: 8 + 2}
  }
}
\hfill
\casepanel{0.285\textwidth}{
  {\bfseries\textcolor{casegreen}{Ours + Chart Parse}}\par
  \vspace{2pt}
  Calls a chart parser and receives structured evidence from the visual content.\par
  \vspace{5pt}
  \caseinnerbox{caseborder}{casecodebg}{\texttt{parse\_content(img1)}}\par
  \vspace{5pt}
  \caseinnerbox{casegreen}{casegreenbg}{
    \centering
    \setlength{\tabcolsep}{4pt}
    \begin{tabular}{lccc}
    \toprule
    \textbf{Country} & \textbf{Ind.} & \textbf{Co.} & \textbf{Total} \\
    \midrule
    China & 12 & 1 & \textbf{13} \\
    Russia & 8 & 2 & \textbf{10} \\
    \bottomrule
    \end{tabular}
  }\par
  \vspace{5pt}
  \caseinnerbox{casegreen}{casegreenbg}{
    {\bfseries\textcolor{casegreen}{Localized evidence:}}\par
    \texttt{China: 12 + 1 = 13}\par
    \texttt{Russia: 8 + 2 = 10}
  }
}
\endgroup
\caption{
Case study of chart evidence acquisition.
The \textsc{CoT} baseline identifies salient bars but fails to recover the corresponding labels and values.
\ours calls \texttt{parse\_content} and obtains structured country-level evidence needed for verification.
}
\label{fig:case_chart_parse}
\end{figure*}

To assess whether the accuracy gains come from better visual evidence acquisition, we sample 100 examples for each evidence type: tables, charts/plots, and multi-panel figures.
For each example, GPT-5.5-assisted annotation identifies the claim-relevant reference evidence, and we report \emph{Relevant Evidence Acquisition Rate} (REAR, defined as the fraction of examples where the model's reasoning trace or tool observations contain the required visual evidence:
$
\operatorname{REAR}
=
\frac{1}{|\mathcal{D}|}
\sum_{x_i\in\mathcal{D}}
\mathbb{I}
\left[
\text{claim-required visual evidence is acquired or identified}
\right].
$
Table~\ref{tab:rear} shows that \ours achieves the highest REAR across all evidence types, raising the average from 83.00\% for \textsc{CoT} and 83.33\% for Tool Prompt to 90.00\%.
The largest gain appears on multi-panel figures, where evidence is often localized and visually dense, while the table subset suggests that prompt-only tool use can retrieve irrelevant evidence.
Together with the qualitative example in Fig.~\ref{fig:case_chart_parse}, these results indicate that learned tool use improves MSCV by helping the model acquire claim-relevant visual evidence.

\subsection{Type-Aware Tool Selection}
\label{sec:type_aware_tool_selection}
We next examine whether, when encouraged to solve MSCV with appropriate tools, the learned policy selects tools according to the visual evidence type.
Using the same diagnostic subsets, we compare Tool Prompt and \ours by accuracy change over non-tool \textsc{CoT}, tool-execution success, and tool-family distribution.

\begin{table*}[t]
\centering
\footnotesize
\renewcommand{\arraystretch}{1.08}
\setlength{\tabcolsep}{5pt}
\begin{tabular}{llccccc}
\toprule
\textbf{Visual Subset} & \textbf{Method} & \textbf{$\Delta$ Acc.} & \textbf{Tool Succ.} & \textbf{$\rho_{\mathrm{tab}}$} & \textbf{$\rho_{\mathrm{chart}}$} & \textbf{$\rho_{\mathrm{zoom}}$} \\
\midrule
Table & Tool Prompt & +3.30 & 88.29 & 78.05 & 20.98 & 0.98 \\
Table & Ours & \textbf{+16.15} & \textbf{97.39} & \textbf{89.25} & 9.77 & 0.98 \\
\midrule
Chart/plot & Tool Prompt & +0.80 & 93.51 & 23.38 & 67.53 & 9.09 \\
Chart/plot & Ours & \textbf{+11.51} & \textbf{96.51} & 21.40 & \textbf{74.24} & 4.37 \\
\midrule
Multi-panel & Tool Prompt & -0.99 & 48.89 & 0.00 & 51.11 & 48.89 \\
Multi-panel & Ours & \textbf{+7.19} & \textbf{50.77} & 4.62 & 41.54 & \textbf{53.85} \\
\bottomrule
\end{tabular}
\caption{
Tool-use behavior across visual-evidence subsets.
$\Delta$ Acc. is the accuracy change in percentage points relative to non-tool \textsc{CoT}; Tool Succ. is the tool-execution success rate; $\rho_{\mathrm{tab}}$, $\rho_{\mathrm{chart}}$, and $\rho_{\mathrm{zoom}}$ are the shares of table, chart, and zoom tool calls.
All remaining values are percentages.
}
\label{tab:tool_use_analysis}
\end{table*}

Table~\ref{tab:tool_use_analysis} shows that \ours learns a more type-aware tool-use policy than Tool Prompt.
On table and chart/plot subsets, \ours routes most calls to the corresponding tool family and obtains much larger accuracy gains.
For multi-panel evidence, \ours makes \texttt{image\_zoom\_in} the largest call family and improves accuracy, whereas Tool Prompt splits calls between chart parsing and zooming and degrades performance.
Thus, GRPO improves not only tool-call executability but also evidence-aligned tool routing.
\subsection{Efficient Tool Use}
\label{sec:efficient_tool_use}
We ablate the OTC efficiency term $s_{\text{eff}}$ while keeping all other reward components and evaluation settings fixed, and evaluate both variants under a tool-encouraging test-time prompt to stress-test whether $s_{\text{eff}}$ reduces redundant tool use and verbosity without sacrificing verification accuracy. We also studied the role of Tool-Interaction Penalty in ~\ref{app:tool_penalty_training_dynamics}.
\begin{table}[t]
\centering
\small
\renewcommand{\arraystretch}{1.08}
\begin{tabular}{lcccc}
\toprule
\textbf{Variant} & \textbf{\textsc{SciVer} Acc.} & \textbf{\textsc{MuSci} Acc.} & \textbf{Calls / Ex.} & \textbf{Avg. Tokens} \\
\midrule
w/o OTC reward $s_{\text{eff}}$ & 76.96 & 73.47 & 1.05 & 2760.42 \\
w/ OTC reward $s_{\text{eff}}$ & \textbf{81.20} & \textbf{78.61} & \textbf{0.74} & \textbf{1361.58} \\
\bottomrule
\end{tabular}
\caption{
Effect of the OTC reward $s_{\text{eff}}$.
Accuracy is reported separately on \textsc{SciVer} and \textsc{MuSciClaims}.
Calls per example and average response length are averaged over both evaluation sets.
}
\label{tab:efficiency_ablation}
\end{table}

Table~\ref{tab:efficiency_ablation} shows that adding $s_{\text{eff}}$ improves both accuracy and efficiency: \textsc{SciVer} and \textsc{MuSciClaims} accuracy increase by 4.24 and 5.14 points, while calls per example drop from 1.05 to 0.74 and average response length decreases from 2760.42 to 1361.58 tokens.
This shows that $s_{\text{eff}}$ regularizes the policy toward shorter and more economical reasoning trajectories rather than simply suppressing tool use.

\subsection{Reasoning over Acquired Evidence}
\label{sec:reasoning_over_evidence}

Beyond visual evidence acquisition and type-aware tool selection (Secs.~\ref{sec:rear_analysis} and~\ref{sec:type_aware_tool_selection}), we ask whether \ours can also reason more reliably once the relevant evidence is provided.
To isolate evidence-based reasoning, we use a text-only evidence-conditioned setting where each input contains the claim, original textual context/caption, and the GPT-5.5-annotated gold visual-evidence statement from Sec.~\ref{sec:rear_analysis}, but excludes images, tool observations, tool schemas, and tool-use instructions.

\begin{wraptable}[9]{r}{0.54\linewidth}
\vspace{-6pt}
\centering
\footnotesize
\setlength{\tabcolsep}{2.6pt}
\renewcommand{\arraystretch}{1.04}
\begin{tabular}{lcccc}
\toprule
\textbf{Model} & \textbf{Overall} & \textbf{Table} & \textbf{Chart} & \textbf{M-panel} \\
\midrule
Qwen3.5-4B & 60.33 & 60.00 & 75.00 & 46.00 \\
\ours & \textbf{71.67} & \textbf{71.00} & \textbf{84.00} & \textbf{60.00} \\
\midrule
$\Delta$ & +11.33 & +11.00 & +9.00 & +14.00 \\
\bottomrule
\end{tabular}
\caption{Evidence-conditioned scientific claim verification accuracy (\%).}
\label{tab:evidence_conditioned}
\vspace{-10pt}
\end{wraptable}

Table~\ref{tab:evidence_conditioned} shows that \ours improves overall accuracy from 60.33\% to 71.67\%, yielding an absolute gain of 11.33 percentage points under the same gold evidence.
The gains are consistent across evidence types, with improvements of 11.00, 9.00, and 14.00 points on table, chart, and multi-panel examples, respectively.
A paired comparison further supports this improvement: \ours corrects 59 examples missed by the base model, while the base model corrects 25 examples missed by \ours, yielding a McNemar exact test result of $p=2.66\times10^{-4}$.
\paragraph{General chart and figure reasoning.}
We further evaluate whether the improvement transfers beyond MSCV on ChartQA~\citep{masry2022chartqa}, DVQA~\citep{kafle2018dvqa}, and FigureQA~\citep{kahou2018figureqa}, using 1{,}000 examples from each benchmark with only the original chart/figure image and question, without gold evidence or tool-use instructions.

\begin{wraptable}[8]{r}{0.50\linewidth}
\vspace{-6pt}
\centering
\footnotesize
\setlength{\tabcolsep}{3.0pt}
\renewcommand{\arraystretch}{1.04}
\begin{tabular}{lccc}
\toprule
\textbf{Benchmark} & \textbf{Qwen3.5} & \textbf{\ours} & \textbf{$\Delta$} \\
\midrule
ChartQA & 63.70 & \textbf{65.40} & +1.70 \\
DVQA & 84.70 & \textbf{90.10} & +5.40 \\
FigureQA & 50.70 & \textbf{59.00} & +8.30 \\
\midrule
Avg. & 66.37 & \textbf{71.50} & +5.13 \\
\bottomrule
\end{tabular}
\caption{External chart/figure reasoning accuracy (\%).}
\label{tab:general_chart_reasoning}
\vspace{-10pt}
\end{wraptable}

Table~\ref{tab:general_chart_reasoning} shows that \ours consistently improves over the backbone, raising average accuracy from 66.37\% to 71.50\% with the largest gains on DVQA and FigureQA.
A paired comparison over all 3{,}000 examples confirms that the improvement is statistically significant ($p=4.66\times10^{-16}$), suggesting that our training also strengthens general chart and figure reasoning beyond the MSCV task.

\section{Conclusion}
We presented \ours, a type-aware visual tool augmented reinforcement learning framework for multimodal scientific claim verification. 
\ours equips VLMs with specialized tools for tables, charts, and general scientific figures, and trains them with GRPO to selectively acquire and use claim-relevant visual evidence. 
Experiments on \textsc{SciVer} and \textsc{MuSciClaims} show that \ours consistently improves over non-tool, prompt-only, and RL-based tool-use baselines. 
These results demonstrate the value of task-aligned visual evidence extraction and learned selective tool use for reliable scientific claim verification.
\section*{Limitation}

ToolSciVer depends on external visual tools to expose structured evidence, so its performance is bounded by their reliability. 
Errors from OCR-based table reading or chart parsing, especially on dense tables, small labels, multi-series plots, or uncommon chart formats, may produce incomplete or incorrect observations and propagate to the verifier. 
Although reinforcement learning improves tool selection and usage, it cannot fully recover from erroneous tool outputs; future work should explore stronger scientific parsers and uncertainty-aware verification mechanisms.


\bibliographystyle{plainnat}
\bibliography{neurips2026_reference}

\appendix
\section{Method Details}
\subsection{Tool Interfaces}
\label{app:tool_interfaces}

Each tool call targets a single image through the argument \texttt{image\_path}, such as \texttt{img1}, \texttt{img2}, or a crop returned by a previous zoom operation.
The scheduler checks that each assistant turn contains either exactly one valid tool call or a final answer.
Malformed calls, unsupported tool names, missing arguments, invalid argument types, or execution failures are rejected and recorded for the tool-interaction penalty.

\paragraph{Table Focus.}
For table images, we first convert the original image into a structured CSV-like representation using OCR.
We then expose two focused tools:
\[
\texttt{focus\_row(image\_path, row\_number)}
\]
and
\[
\texttt{focus\_column(image\_path, column\_number)}.
\]
The row tool returns the content of the selected row, and the column tool returns the content of the selected column.
This design is intended to expose localized tabular evidence, such as a specific method row, dataset row, metric column, or numerical entry, without returning the entire table.

\paragraph{Chart Parse.}
For chart and plot images, we expose:
\[
\texttt{parse\_content(image\_path)}.
\]
The tool returns parsed chart content as structured text when possible.
For example, the returned observation may contain chart elements, axis labels, series names, and associated values in a JSON-like format.
When strict structural parsing is unavailable, the tool falls back to raw extracted content rather than failing silently.
This allows the verifier to compare a claim against explicit chart evidence such as trends, thresholds, relative ordering, or numerical values.

\paragraph{Region Zoom.}
For general scientific figures, diagrams, dense images, or multi-panel figures, we expose:
\[
\texttt{image\_zoom\_in(image\_path, }x_1\texttt{, }y_1\texttt{, }x_2\texttt{, }y_2\texttt{)}.
\]
The tool crops and enlarges the selected region and returns the zoomed-in image.
The bounding box uses normalized coordinates on a $[0,1000]$ scale with respect to the selected image.
This makes the interface resolution-invariant and allows the model to inspect small local regions that may contain decisive visual evidence.

\subsection{Tool Scheduler and Execution Logs}
\label{app:tool_scheduler}

All tool-use methods interact with tools through the same scheduler.
The scheduler parses model outputs, validates tool-call syntax, executes valid calls, and returns either a tool observation or an execution error.
Each assistant turn is constrained to contain either one valid tool call or the final answer.
If a turn contains multiple tool calls, an unsupported tool name, missing arguments, invalid argument types, coordinates outside the valid range, or an unparsable call format, the scheduler records it as a malformed tool turn.

The scheduler records the following metadata for each rollout:
tool name, tool arguments, execution success, execution failure type, malformed-call indicator, number of tool calls, response length, and final-answer validity.
These logs are used for both training and analysis.
During training, malformed calls and execution failures contribute to the tool-interaction penalty.
During evaluation, the logs are used to compute tool success rate, tool-family routing statistics, tool calls per example, and tool calls per correct example.

\subsection{GRPO Objective and Reward Details}
\label{app:grpo_reward_details}

\paragraph{GRPO objective.}
For each input instance $x$, we sample a group of $G$ rollouts
\[
\mathcal{G}(x)=\{\tau_i\}_{i=1}^{G}
\]
from the current policy.
Each rollout may contain zero or more tool calls before the final answer.
After assigning rollout-level rewards $R_i$, we compute the group-relative advantage:
\[
A_i =
\frac{R_i - \operatorname{mean}_{j=1}^{G} R_j}
{\operatorname{std}_{j=1}^{G} R_j}.
\]
The policy is optimized with the clipped GRPO objective:
\[
\mathcal{J}_{\text{GRPO}}(\theta)
=
\frac{1}{G}
\sum_{i=1}^{G}
\sum_{t}
\min
\left(
r_{i,t}(\theta) A_i,
\operatorname{clip}(r_{i,t}(\theta),1-\epsilon,1+\epsilon)A_i
\right)
-
\beta D_{\mathrm{KL}}(\pi_{\theta}\|\pi_{\mathrm{ref}})
\]
where $r_{i,t}(\theta)$ is the token-level policy ratio, $\pi_{\mathrm{ref}}$ is the reference policy, and $\beta$ controls the KL penalty.

\paragraph{Overall reward.}
For a rollout $\tau$, the final reward is
\[
R(\tau)
=
s_{\text{eff}}(\tau)
\left(
r_{\text{ans}}(\tau)
+
\lambda_{\text{fmt}} r_{\text{fmt}}(\tau)
\right)
+
\lambda_{\text{len}} r_{\text{len}}(\tau)
+
\lambda_{\text{tool}} r_{\text{tool}}(\tau).
\]
In our experiments, we set
\[
\lambda_{\text{fmt}}=\lambda_{\text{len}}=\lambda_{\text{tool}}=0.1.
\]
The answer and format rewards form the task-critical part of the objective, while the length and tool-interaction terms discourage overlong generations and invalid tool behavior.
The coefficient $s_{\text{eff}}$ encourages efficient tool use relative to the best correct trajectories in the sampled GRPO group.

\paragraph{Answer reward.}
The answer reward measures final task correctness:
\[
r_{\text{ans}}(\tau)=\mathbb{I}[\hat{y}=y],
\]
where $y$ is the ground-truth label and $\hat{y}$ is the normalized model prediction.

\paragraph{Format reward.}
The format reward encourages the model to produce a parseable final response.
We define
\[
r_{\text{fmt}}(\tau)
=
0.5\,I_{\text{reason}}(\tau)
+
0.5\,I_{\text{answer}}(\tau),
\]
where $I_{\text{reason}}(\tau)$ indicates whether the final response contains the required reasoning segment under the prompt template, and $I_{\text{answer}}(\tau)$ indicates whether the response contains exactly one valid normalized final-answer marker.
This term stabilizes training and ensures that final predictions can be reliably extracted.

\paragraph{Tool-efficiency coefficient.}
To encourage selective rather than excessive tool use, we introduce an OTC-style tool-efficiency coefficient.
Let $m(\tau)$ be the number of tool calls in rollout $\tau$.
Let $\mathcal{C}(x)$ be the set of strict-correct rollouts in the GRPO group, where a rollout is strict-correct if it predicts the correct label and ends with a valid final answer.
If $\mathcal{C}(x)$ is non-empty, we estimate the group-local optimal tool budget as
\[
n^\star(x)=\min_{\tau_i\in\mathcal{C}(x)} m(\tau_i).
\]
If no strict-correct rollout exists, we set $s_{\text{eff}}(\tau)=1$.
Otherwise, for maximum tool budget $c_{\max}$, we define
\[
s_{\text{eff}}(\tau)=
\begin{cases}
1, & n^\star=0,\; m=0, \\
\cos\left(\frac{m\pi}{2m+c_{\max}}\right),
& n^\star=0,\; m>0, \\
0, & n^\star>0,\; m=0, \\
\sin\left(\frac{m\pi}{m+n^\star}\right),
& n^\star>0,\; m>0.
\end{cases}
\]
This coefficient rewards correct trajectories that use close to the group-local minimal number of tool calls, while penalizing unnecessary tool use when a direct answer is sufficient.
We set $c_{\max}=5$.

\paragraph{Length reward.}
We use a soft overlong penalty to discourage unnecessarily long generations.
Let $L(\tau)$ denote the completion length in tokens, let $L_{\max}$ be the maximum completion length, and let $C$ denote the soft-cache length.
We define the soft threshold as
\[
L_{\text{soft}}=L_{\max}-C.
\]
The length reward is
\[
r_{\text{len}}(\tau)
=
-\frac{\max(L(\tau)-L_{\text{soft}},0)}{C}.
\]
Thus, the penalty is zero below the soft threshold and decreases linearly as the response approaches the hard length limit.
In our experiments, $L_{\max}=4096$ and $C=768$.

\paragraph{Tool-interaction penalty.}
The tool-interaction penalty discourages malformed tool turns and execution failures.
Let $e_{\text{turn}}(\tau)$ be the number of malformed tool turns, let $e_{\text{exec}}(\tau)$ be the number of tool-execution failures, and let $I_{\text{bad-end}}(\tau)$ indicate whether the rollout contains at least one tool-related error and still fails to terminate with a valid final answer.
We compute
\[
\begin{aligned}
\tilde{r}_{\text{tool}}(\tau)=&
-0.20\,\mathbb{I}[e_{\text{turn}}\ge 1]
-0.10\,\max(e_{\text{turn}}-1,0) \\
&-0.30\,\mathbb{I}[e_{\text{exec}}\ge 1]
-0.15\,\max(e_{\text{exec}}-1,0) \\
&-0.40\,I_{\text{bad-end}}(\tau),
\end{aligned}
\]
and clip the result:
\[
r_{\text{tool}}(\tau)
=
\operatorname{clip}_{[-1.5,0]}
\left(\tilde{r}_{\text{tool}}(\tau)\right).
\]
This gives the model dense feedback for invalid tool behavior, rather than treating all tool failures as a single sparse error.

\section{Experiment Details}
\label{app:additional_details}

\subsection{Training and Test Set Distribution}
\label{app:data_distribution}

Table~\ref{tab:appendix_train_distribution} and Table~\ref{tab:appendix_test_distribution} summarize the exact split sizes used in our experiments.
We use a mixed training set with 3,010 examples in total.
The evaluation reported in the main text uses 2,505 test instances in total: 2,000 from \textsc{SciVer} and 505 from the \textsc{MuSciClaims} test split.

These counts are also the reason we report \emph{sample-weighted} overall accuracy in the main results table.
\textsc{SciVer} is nearly balanced across its four reasoning categories, but the current \textsc{MuSciClaims} split is not balanced across biology, physics, and chemistry.
A simple macro average would therefore over-weight the smaller physics and chemistry subsets.

\begin{table}[h]
\centering
\small
\renewcommand{\arraystretch}{1.06}
\begin{tabular}{llr}
\toprule
\textbf{Split} & \textbf{Category} & \textbf{\# Examples} \\
\midrule
\textsc{SciVer} train mixture & Analytical & 494 \\
 & Direct & 502 \\
 & Parallel & 505 \\
 & Sequential & 499 \\
 & Subtotal & 2,000 \\
\midrule
\textsc{MuSciClaims} train mixture & Biology & 612 \\
 & Physics & 192 \\
 & Chemistry & 206 \\
 & Subtotal & 1,010 \\
\midrule
\textbf{Total training mixture} & -- & \textbf{3,010} \\
\bottomrule
\end{tabular}
\caption{Category distribution of the mixed training set used for the GRPO runs.}
\label{tab:appendix_train_distribution}
\end{table}

\begin{table}[h]
\centering
\small
\renewcommand{\arraystretch}{1.06}
\begin{tabular}{llr}
\toprule
\textbf{Split} & \textbf{Category} & \textbf{\# Examples} \\
\midrule
\textsc{SciVer} test & Analytical & 494 \\
 & Direct & 502 \\
 & Parallel & 505 \\
 & Sequential & 499 \\
 & Subtotal & 2,000 \\
\midrule
\textsc{MuSciClaims} test & Biology & 306 \\
 & Physics & 96 \\
 & Chemistry & 103 \\
 & Subtotal & 505 \\
\midrule
\textbf{Total evaluation set} & -- & \textbf{2,505} \\
\bottomrule
\end{tabular}
\caption{
Category distribution of the test splits used in the current experiments.
The main-table ``Overall'' values are computed by weighting each category by these test-set counts.
}
\label{tab:appendix_test_distribution}
\end{table}

\subsection{Training and Evaluation Protocol.}
For each trainable method, we fine-tune the corresponding backbone on the mixed MSCV training set.
During training and evaluation, a model may either produce a final answer directly or issue tool calls through the scheduler.
Each assistant turn contains at most one tool call, and the total tool budget is capped at $c_{\max}=5$.
The scheduler validates tool-call syntax, executes valid calls, returns tool observations or execution errors, and records tool-use metadata for later analysis.

All models are evaluated in a closed-context setting using only the benchmark-provided claim, visual input, caption, and textual context.
We do not allow open-web retrieval or external corpora at evaluation time.
Free-form generations are normalized to the benchmark label space with rule-based answer extraction, and accuracy is the primary metric.
For diagnostic analyses, we additionally report Relevant Evidence Acquisition Rate (REAR), type-aware tool routing statistics, tool calls per example, tool calls per correct example, and response length; metric definitions are provided in Appendix~\ref{app:diagnostic_metrics}.

\subsection{Baseline and Prompting Details}
\label{app:baseline_prompting}

\paragraph{Non-tool CoT inference.}
The non-tool CoT baseline receives the claim, context, caption, and image, but has no access to external visual tools.
It is prompted to reason step by step and then produce a normalized yes/no answer.
The prompt template is:

\begin{verbatim}
Claim: $claim
Context: $context
Caption: $caption

Your task is to critically evaluate the claim based on the image and the caption.
Carefully examine whether the information in the caption truly supports the claim.
Be skeptical and cautious: if there is any inconsistency, missing evidence, or ambiguity,
consider the claim incorrect.

Start by explaining your reasoning process clearly, focusing on identifying potential
contradictions, lack of support, or misleading interpretations. If the claim is unsupported
or contradicted by the caption and image, respond with 'no'. Only respond with 'yes'
if the claim is fully and clearly supported.

Conclude your analysis by stating: 'Therefore, the final answer is: Answer: $$ANSWER',
where $$ANSWER is your final answer. Think step by step before answering.
\end{verbatim}

\paragraph{Prompt-only tool-use CoT inference.}
The prompt-only tool-use baseline receives the same MSCV input and the same tool APIs as our method.
However, the model is not trained with tool-use trajectories or reward feedback.
It must decide from prompting alone whether a tool is needed, which tool to call, and what arguments to provide.
This baseline isolates the effect of exposing tools without learning a tool-use policy.

\paragraph{Dataset-adapted trained tool-use baselines.}
For the \textsc{VTool-R1} and \textsc{OpenThinkIMG} baselines, we retain their original tool suites and reward designs.
We train each baseline on the same MSCV training splits used by our method.
This comparison tests whether general visual tool-use training recipes are sufficient for MSCV, or whether task-aligned evidence extraction tools are needed.

\paragraph{Proprietary model references.}
Proprietary VLMs are evaluated only under the non-tool CoT inference setting.
They are included as external references rather than controlled training baselines, since their training data, model parameters, and tool-use training procedures are not available.

\subsection{Diagnostic Subset Construction}
\label{app:diagnostic_subset_construction}

We construct diagnostic subsets to analyze whether the model acquires the correct type of visual evidence.
These subsets are used for the REAR analysis and for the type-aware tool-selection analysis in Section~\ref{sec:results}.

\paragraph{Table subset.}
The table subset is drawn from \textsc{SciVer} examples whose evidence images contain table indicators in the file name or metadata.
These examples are used to evaluate whether the model can acquire localized tabular evidence, such as a target row, column, or cell value.

\paragraph{Chart/plot subset.}
The chart/plot subset is drawn from \textsc{SciVer} examples for which chart pre-computation metadata indicates that chart parsing is applicable.
These examples test whether the model can acquire chart evidence such as trends, thresholds, relative ordering, or numerical values.

\paragraph{General-figure subset.}
The general-figure subset is drawn from \textsc{MuSciClaims}, whose examples are centered on figure-centric scientific claims.
This subset is used to evaluate whether the model can acquire local evidence from dense, diagrammatic, or multi-panel scientific figures.
When a claim requires multiple evidence types, an example may contribute to more than one diagnostic subset.

\subsection{Evaluation Metrics for Diagnostic Analyses}
\label{app:diagnostic_metrics}

In addition to final verification accuracy, we compute diagnostic metrics for the analyses in Section~\ref{sec:results}.
These metrics are designed to test whether the model improves because it acquires claim-relevant visual evidence, selects tools according to the visual evidence type, and solves the task efficiently.

\paragraph{Sample-weighted overall accuracy.}
For each benchmark, the overall accuracy is computed by weighting each subset by its number of test examples:
\[
\operatorname{Acc}_{\text{overall}}
=
\frac{
\sum_{k} n_k \operatorname{Acc}_k
}{
\sum_{k} n_k
},
\]
where $k$ indexes benchmark subsets, $n_k$ is the number of examples in subset $k$, and $\operatorname{Acc}_k$ is the accuracy on that subset.

\paragraph{Relevant Evidence Acquisition Rate.}
Relevant Evidence Acquisition Rate (REAR) measures whether a model obtains the visual evidence required to verify the claim.
For tool-use methods, an example is counted as successful if at least one returned tool observation contains the claim-relevant visual evidence.
For non-tool CoT inference, we compute a rationale-level REAR by checking whether the generated reasoning explicitly identifies the required visual evidence.
Formally, for a diagnostic subset $\mathcal{D}$,
\[
\operatorname{REAR}
=
\frac{1}{|\mathcal{D}|}
\sum_{x_i\in\mathcal{D}}
\mathbb{I}[
\text{claim-relevant visual evidence is acquired or identified}
].
\]
We report REAR separately for table, chart/plot, and general-figure evidence subsets.


\paragraph{Tool success rate.}
Tool success rate measures the fraction of attempted tool calls that execute successfully:
\[
\operatorname{ToolSucc}
=
\frac{
\#\text{ successfully executed tool calls}
}{
\#\text{ attempted tool calls}
}.
\]
A low tool success rate indicates malformed calls, invalid arguments, unsupported tool names, or execution failures.

\paragraph{Tool-family routing share.}
For a tool family $f$, the routing share is
\[
\rho_f
=
\frac{
\#\text{ calls to tool family } f
}{
\#\text{ all valid tool calls}
}.
\]
In the main analysis, we aggregate \texttt{focus\_row} and \texttt{focus\_column} into the table-tool family, \texttt{parse\_content} into the chart-tool family, and \texttt{image\_zoom\_in} into the zoom-tool family.

\paragraph{Tool-use efficiency.}
We measure tool-use efficiency using the average number of tool calls per example and the average number of tool calls per correct example:
\[
\operatorname{Calls/Ex}
=
\frac{1}{N}
\sum_{i=1}^{N}
m(\tau_i),
\]
\[
\operatorname{Calls/Correct}
=
\frac{
\sum_{i=1}^{N} m(\tau_i)\,\mathbb{I}[\hat{y}_i=y_i]
}{
\sum_{i=1}^{N}\mathbb{I}[\hat{y}_i=y_i]
}.
\]
We also report average response length in tokens.
Together, these metrics test whether the learned policy solves MSCV with fewer unnecessary actions and shorter responses while preserving accuracy.

\subsection{Training Dynamics with Tool-Interaction Penalty}
\label{app:tool_penalty_training_dynamics}

\begin{figure}[t]
\centering
\includegraphics[width=\linewidth]{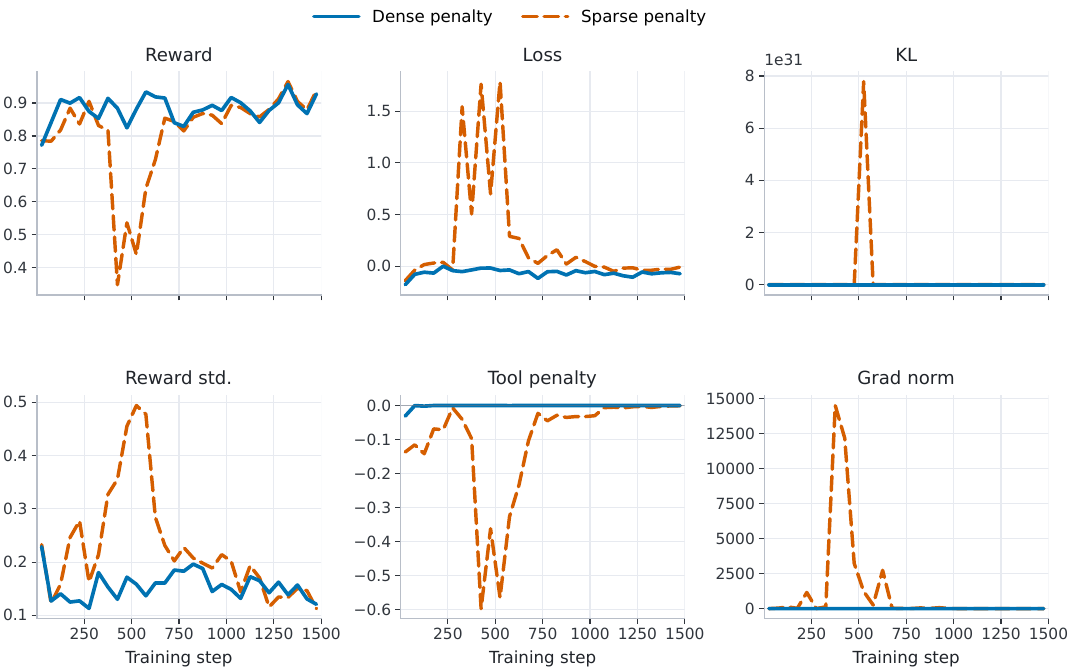}
\caption{
Training dynamics for sparse versus dense tool-interaction penalties.
Curves are smoothed by averaging metrics within 50-step intervals.
The dense tool-interaction penalty produces smoother reward, loss, KL, reward-variance, tool-penalty, and gradient-norm trajectories.
}
\label{fig:tool_penalty_training_dynamics}
\end{figure}

We additionally analyze how the tool-interaction penalty affects GRPO training dynamics.
Figure~\ref{fig:tool_penalty_training_dynamics} compares a sparse tool-penalty variant with the dense tool-interaction penalty used in our method.
Both runs use the same backbone, training data, reward weights, and GRPO hyperparameters; they differ only in how invalid tool behavior is penalized.

The sparse variant provides coarse feedback for tool-related failures, whereas the dense variant assigns graded penalties to malformed tool turns, execution failures, repeated errors, and invalid termination after tool errors.
The dense penalty leads to smoother optimization behavior across reward, loss, KL, reward variance, tool-penalty reward, and gradient norm.
This suggests that fine-grained feedback makes tool-use learning more stable by reducing abrupt reward changes caused by sparse failure signals.

\end{document}